\documentclass[10pt,twocolumn,letterpaper]{article}

\usepackage[pagenumbers]{cvpr} 

\usepackage{graphicx}
\usepackage{amsmath}
\usepackage{amssymb}
\usepackage{booktabs}
\usepackage{algorithm}
\usepackage{algpseudocode}
\usepackage{multirow}
\usepackage{bbm}
\usepackage[accsupp]{axessibility} 

\usepackage[pagebackref,breaklinks,colorlinks]{hyperref}

\usepackage[capitalize]{cleveref}
\crefname{section}{Sec.}{Secs.}
\Crefname{section}{Section}{Sections}
\Crefname{table}{Table}{Tables}
\crefname{table}{Tab.}{Tabs.}

\def\cvprPaperID{*****} 
\def\confName{CVPR}
\def\confYear{2023}

\begin{document}

\title{Systematic Architectural Design of Scale Transformed Attention Condenser DNNs via Multi-Scale Class Representational Response Similarity Analysis}

\author{Andrew Hryniowski$^{1,2,3}$, Alexander Wong$^{1,2,3}$\\
			$^{1}$ Vision and Image Processing Research Group, University of Waterloo\\
			$^{2}$ Waterloo Artificial Intelligence Institute, Waterloo, ON\\
			$^{3}$ DarwinAI Corp., Waterloo, ON\\ 			
			\texttt{$\{$apphryni, a28wong$\}$@uwaterloo.ca}
	}

\maketitle

\begin{abstract}

Self-attention mechanisms are commonly included in a convolutional neural networks to achieve an improved efficiency performance balance. However, adding self-attention mechanisms adds additional hyperparameters to tune for the application at hand. In this work we propose a novel type of DNN analysis called Multi-Scale Class Representational Response Similarity Analysis (ClassRepSim) which can be used to identify specific design interventions that lead to more efficient self-attention convolutional neural network architectures. Using insights grained from ClassRepSim we propose the Spatial Transformed Attention Condenser (STAC) module, a novel attention-condenser based self-attention module. We show that adding STAC modules to ResNet style architectures can result in up to a $1.6\%$ increase in top-1 accuracy compared to vanilla ResNet models and up to a $0.5\%$ increase in top-1 accuracy compared to SENet models on the ImageNet64x64 dataset, at the cost of up to $1.7\%$ increase in FLOPs and 2x the number of parameters. In addition, we demonstrate that results from ClassRepSim analysis can be used to select an effective parameterization of the STAC module resulting in competitive performance compared to an extensive parameter search. 

\end{abstract}

\begin{figure}
    \centering
    \includegraphics[width=0.9\linewidth]{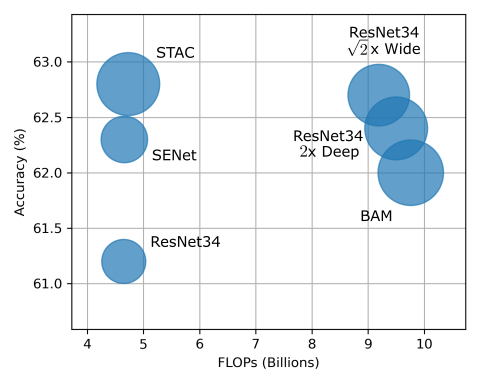}
    \caption{A comparison between six ResNet34 based architectures trained on the ImageNet64x64 dataset. The proposed STAC based model achieves the highest performance while requiring a fraction of FLOPs and a similar number of parameters (proportional to the associated circle area) compared to the next closest model. The precise numbers can be found in Table~\ref{tab:results}. The STAC, SENet, and BAM models in this figure use standard module placement.}
    \label{fig:stac_module_bubble}
\end{figure}

\section{Introduction}
\label{sec:intro}
Designing targeted deep neural network (DNN) architectures is a common method for achieving desired trade-offs between performance and efficiency requirements. Adding self-attention modules can be an efficient method for improving the performance-efficiency trade-offs of a deep neural network (DNN)~\cite{wong2020tinyspeech}. Knowing which self-attention module to use is often hard to do determine and can require extensive hyperparameter testing. 

In this paper, we propose a novel form of analysis which studies the relationship between samples of a dataset and how it changes throughout a DNN. We call this analysis Multi-Scale Class Representational Response Similarity Analysis (ClassRepSim). Our study focuses on whether neighbouring samples in a model's latent space are of the same class. We use this information to study how the class similarity (CS) changes between layers in a DNN, and how the relationship between samples changes when view from different spatial resolutions. By observing these changes we reveal novel aspects in the learned latent representation of ResNet models. 

\begin{figure*}
    \centering
    \includegraphics[width=0.7\textwidth]{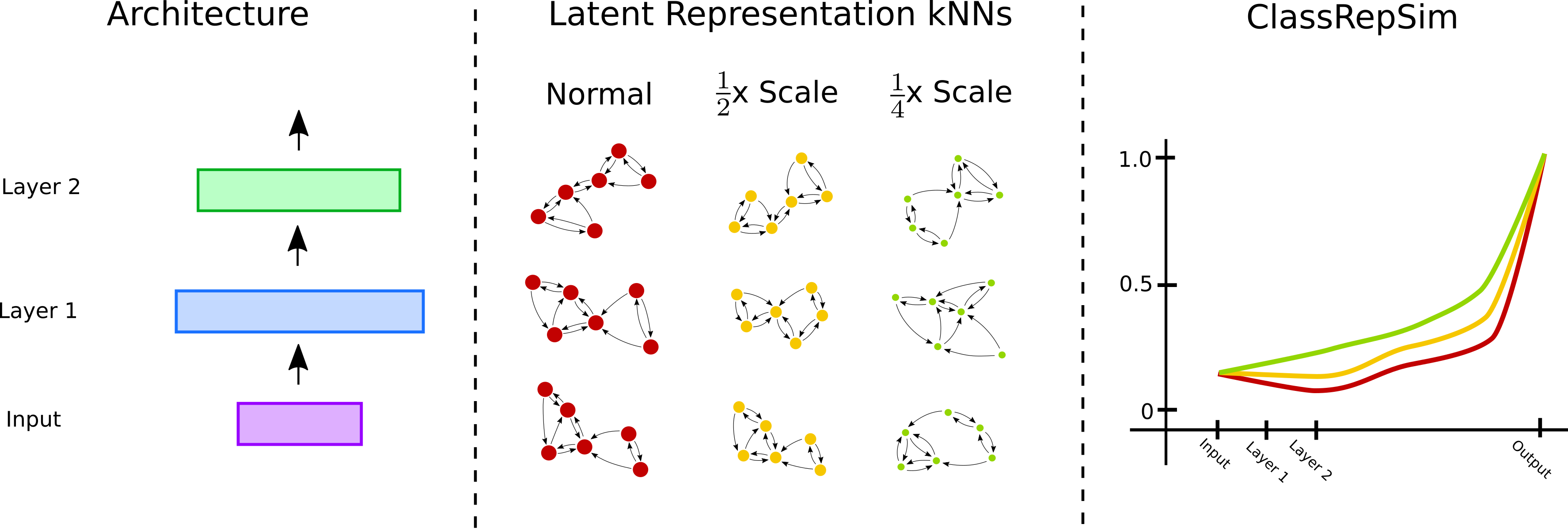}
    \caption{An example of the ClassRepSim proccess. At each layer latent features for each sample are extracted. Features for a layer are used to calculate a kNN graph. This is repeated for each spatial scale under consideration. For each kNN the average class similarity (CS) between samples is calculated. On the right is an example of a set of CS curves, with each curve representing a different resolution.}
    \label{fig:class_sim_resnet20_cifar10}
\end{figure*}

Three key observations in this work are that there exists a optimal scale at which samples are most class-wise similar to one another, that the peak scale at which samples are most class-wise similar changes throughout a model, and that the spatial resolution of the feature maps plays a key role in determining the peak resolutions. Using these insights we propose spatial transformed attention condenser (STAC) modules, an efficient self-attention module which demonstrates a superior FLOPs to performance tradeoff compared to standard ResNet models and ResNet models using existing self-attention module designs. The following are the key contributions of this work:
\begin{enumerate}
    \item \textbf{ClassRepSim}: a novel type of representational response analysis that identifies the class-wise importance of features at varying spatial scales across the layers of a DNN (Section~\ref{sec:class_similarity}), 
    \item \textbf{Scale Transformed Attention Condenser (STAC) Modules}: a novel instantiation of attention condenser modules designed to emphasise the class similarity of the latent features (Section~\ref{sec:spatial_attention_condensers},~\ref{sec:results}), and  
    \item \textbf{STAC Versatility}: A systematic evaluation of the proposed STAC module demonstrating that it achieves competitive performance under a variety of hyperparameter configurations (Section~\ref{sec:ablation}). 
\end{enumerate}

\section{Background}
\label{sec:background}

\textbf{Representational Response:} Analytical methods have long been used to understand the relationship between a DNN and the data distribution used to train it. Methods such as CCA~\cite{hardoon2004canonical}, SVCCA~\cite{raghu2017svcca}, and CKA~\cite{kornblith2019similarity} consider the relative distance between samples in one layer to the distance between the same samples in another layer. Simple to compute methods such as linear probing~\cite{alain2016understanding} give indirect evidence of the similarity between samples by directly learning a linear classifier at each layer of a DNN. Other non-linear methods use kNNs as a basis for calculating similarity metrics. Methods such as Hierarchical Nucleation~\cite{doimo2020hierarchical} use kNNs to study how the latent representations of samples gradually cluster with like samples as the layers get progressively deeper. Using a similar construction to Hierarchical Nucleation, (Hryniowski et al.)~\cite{hryniowski2020inter} uses kNNs to study the similarity of latent representations using Nearest Neighbour Topological Similarity, a method similar to CKA, and Nearest Neighbour Topological Similarity to directly study how layers in a DNN effect the creation and removal of neighbouring samples when comparing kNNs at different points in a DNN. In this work we expand upon the kNN graph approach used in (Hryniowski et al.)~\cite{hryniowski2020inter} to explore the change in class similarity as the spatial resolution of the latent representations are progressively pooled to smaller sizes.    

\textbf{Generative Architectures:} Methods that use metrics to directly generate new model designs include Generative Synthesis~\cite{wong2018ferminets}, Morphnet~\cite{gordon2018morphnet}, and Lotto Ticket~\cite{frankle2018lottery} in which latent representations of data are used to quantify the importance of specific architectural components and remove the components which do not benefit the models performance. KNN based methods like Deep k-nearest neighbour~\cite{papernot2018deep} allow a model's prediction, confidence, credibility to be calculated and compared across models. Other approaches implicitly integrate soft kNNs into the training procedure as a regularizer. The work (Frosst et al.)~\cite{frosst2019analyzing} uses soft nearest neighbour loss~\cite{salakhutdinov2007learning} to maximize the entanglement between classes; counter intuitively the increased class entanglement improves model robustness to adversarial attacks. Within this work we use insights from our proposed analysis to generate a novel architectural module.

\textbf{Architectual Tradeoffs:} Many innovations in DNN design stem from purpose build architectural changes that improve a model's ability to learn information embedded within a dataset. The most widely used and general purpose architectural improvement is the residual skip connection design~\cite{he2016deep}, which allow substantially deeper and more performant models to be trained. Methods such as atrous convolution~\cite{chen2017deeplab} and GoogleLeNet~\cite{szegedy2015going} use different techniques to embed a greater receptive field of a given layer. Techniques like depthwise convolutions~\cite{chollet2017xception}, weight quantization~\cite{wu2016quantized}, and bottleneck~\cite{he2016deep} all make targeted changes and tradeoffs which balance a model's computational efficiency with its performance. The architectural module proposed in this work both increases the receptive field of the layers it is applied too, improves the computational efficiency of a model by preserving model performance while allowing a smaller model to be used, and who's specific location and parameterization is dictated from insights gained using ClassRepSim.

\textbf{Spatial Self-Attention Modules:} Attention models use a specific style of building block that allows a model to more easily focus in on a specific subset of features. Squeeze and excitation network (SENet)~\cite{hu2018squeeze} based designs introduces an attention module that allows an easy method of highlighting specific channels of a feature maps using global feature context. With the use of minimal additional computation the SENet module improves network performance on the ImageNet dataset~\cite{deng2009imagenet} on all base architectures it is applied too. Several derivative methods, including Bottleneck Attention Module (BAM)~\cite{park2018bam} and Spatial and Channel-wise CNNs (SCA-CNNs), expand on the SENet module by adding a separate spatial attention mechanism in a parallel and sequential manner, respectively. Attention Condensers (ACs)~\cite{wong2020tinyspeech} unify these approaches in a common framework that combines channel attention, spatial attention, and focusing through bottlenecks into a single module design. The specific parameterization of AC modules is determined using Generative Synthesis~\cite{wong2018ferminets} based design exploration. In this work the novel module proposed is a specific instantiation of an AC module. However the proposed module design is not obtained through use of a computationally expensive iterative DNN design exploration techniques, but instead obtained directly thought the insights generated by the proposed ClassRepSim analysis.

\section{ClassRepSim: Multi-Scale Class Representational Response Similarity Analysis}
\label{sec:class_similarity}

We briefly describe the kNN approach described in  (Hryniowski et al.)~\cite{hryniowski2020inter} which we use as the foundation for multi scale class representational response similarity analysis. Let $X$ be a set of data with $N$ samples $x \in X$. Let $F$ represent some deep neural network consisting of a set of functions $\{f^i\}$. Let a given function $f^i$ be defined by
\begin{equation}
    Y^i = f^i(Y^j; \theta^i)    
\end{equation}
where $Y^i$ and $Y^j$ are the latent representation of dataset $X$ as seen at the output of operation $f^i$ and $f^j$, respectively. For convenience, let $Y^0 = f^0(X) = X$. For a given set of latent representations $Y^k$, let a directed kNN graph with $M$ neighbours per sample be defined as $g^k = g(Y^k, D^k)$ where $D^k$ is the set of directed edges between vertices $Y^k$. Let the set of neighbours for sample $y^k_a \in Y^k$ in the kNN $g^k$ be $\{h_m\}_{0 < m < M}$. In this work we use these kNN graphs to compare the class similarity between neighbouring samples at different spatial scales. 

\begin{figure*}
    \centering
    \begin{tabular}{cc}
    \subfloat[Sample CS Transitions\label{subfig:cs_toy_example_vis}]{\includegraphics[width=0.6\linewidth]{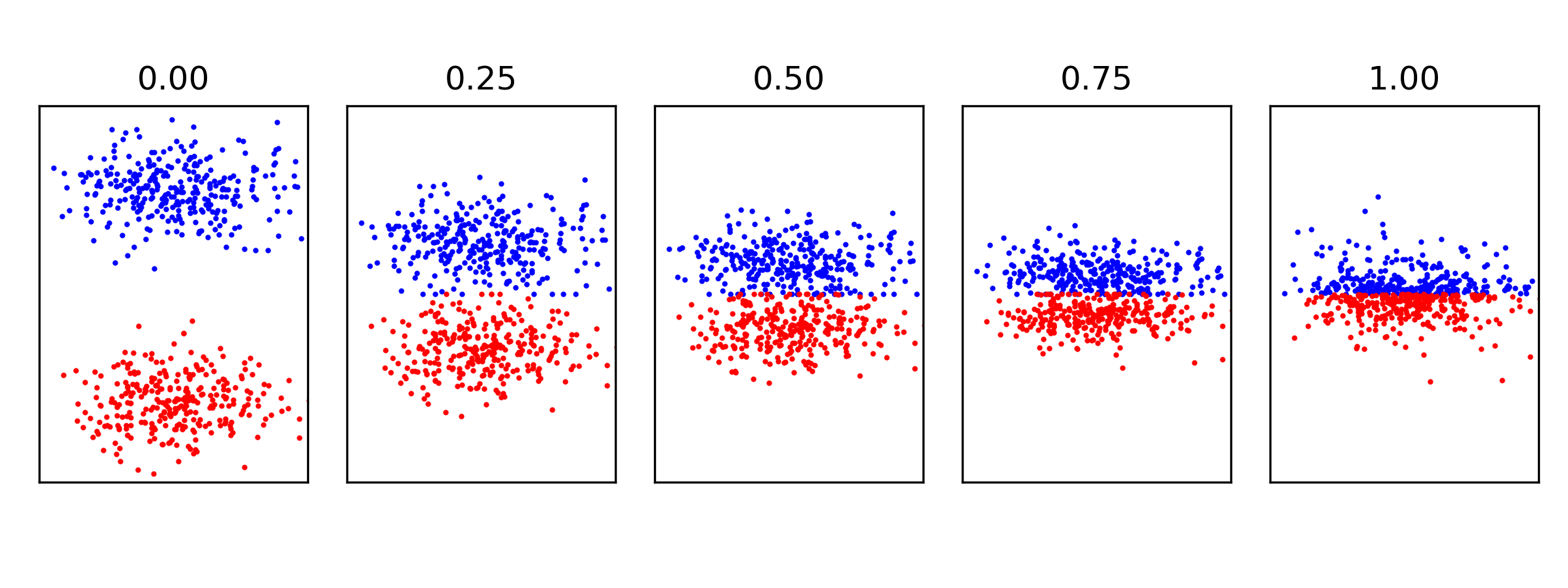}} & 

    \subfloat[CS Transition Curve\label{subfig:cs_toy_example}]{\includegraphics[width=0.25\linewidth]{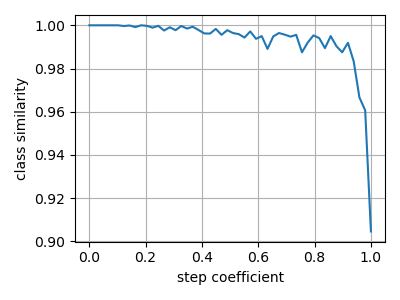}} 
    
    \end{tabular}
    \caption{\small An example of how the class similarity is effected as the distance between two linearly separable class clusters changes. Figure~\ref{subfig:cs_toy_example_vis} shows the transition as two linearly separable classes approach one another. Note that at a transition state of 1.0 the two classes are still linearly separable along the x-axis. Figure~\ref{subfig:cs_toy_example} shows the observed CS for the toy dataset as the transition state moves from 0.0 to 1.0. When the transition state is 0.0 the CS equals 1.0, by the time the transition state reaches 1.0 the CS has reduced to almost 0.9.}
 
    \label{fig:cs_toy_example_group}
\end{figure*}

We now introduce multi-scale class representational response similarity analysis (ClassRepSim). At a high level the proposed method investigates how the class similarity between latent representation a different spatial scales changes. More specifically, for a given set of data $X$ we first extract latent representations of $X$ for each operation of interest in the DNN; in practice we focus primarily on the output of each layer. Each latent representation $Y^i$ is independently spatially down sampled for each scale $s \in S$. Let $Y^i_s = P(Y^i, s)$ be the scaled latent representation of $X$ output by operation $f^i$ and down sampled by scale $s$ using some function $P$. Then for each $Y^i_s$ a kNN $g^i_s$ is calculated and used the calculate the average class similarity $c^i_s$ for the given operation $f^i$ and scale $s$. For any given kNN let the class similarity score $c$ be defined by
\begin{equation}
    c = CS(g, Z) = \frac{1}{MN}\sum_n \sum_m \mathbbm{1}(Z_n, Z_m)
\end{equation}
where $\mathbbm{1}(\cdot)$ is an indicator function, and $Z_n$ and $Z_k$ are the ground truth classes for the $n^{th}$ sample in $N$ and the $m^{th}$ nearest neighbour of the $n^{th}$ sample, respectively. The complete process is described in Algorithm \ref{algo:ClassRepSim}.

\begin{algorithm}
	\caption{ClassRepSim} 
        \hspace*{\algorithmicindent} \textbf{Input}: $X$, $Z$, $F$ \\ 
        \hspace*{\algorithmicindent} \textbf{Output}: $\{c^i_s\}$
	\begin{algorithmic}[1]
		\For {$i$} \Comment{For each operation in DNN $F$ of interest}
                \State $Y^i \gets f^i(X)$  \Comment{Extract features at operation $i$}
			\For {$s$} \Comment{For each scale}
				\State $Y^i_s \gets P(Y^i, s)$ 
                        \Comment{Scale the features}
				\State $g^i_s \gets g(Y^i_s)$   
                        \Comment{Calculate the kNN}
                    \State $c^i_s \gets CS (g^i_s, Z)$          
                        \Comment{Calculate class similarity}
			\EndFor
		\EndFor
	\end{algorithmic} 
\label{algo:ClassRepSim}
\end{algorithm}

\subsection{Models}
\label{sec:model}
In this section we use a ResNet20 model~\cite{he2016deep} using the CIFAR style block design and post batchnorm activation. We use two datasets: CIFAR10 and a 50 class subset of ImageNet64x64~\cite{chrabaszcz2017downsampled} which we refer to as Img64-50. We use SGD with momentum, a base learning of $0.01$, a cosine decay rate that ends at $1\%$ the initial learning rate, weight decay of $1e-4$, training for 100 epochs, and warm up of 10 epochs. All results in this work are the average of 8 models.

\subsection{Multi-Scale Class Similarity}
\label{sec:multi-scale-cs}

\begin{figure}[ht]
    \centering
    \begin{tabular}{cc}
    \subfloat[CIFAR10 \label{subfig:feature_dim_avgpool_CS_cifar10}]{\includegraphics[width=0.45\linewidth]{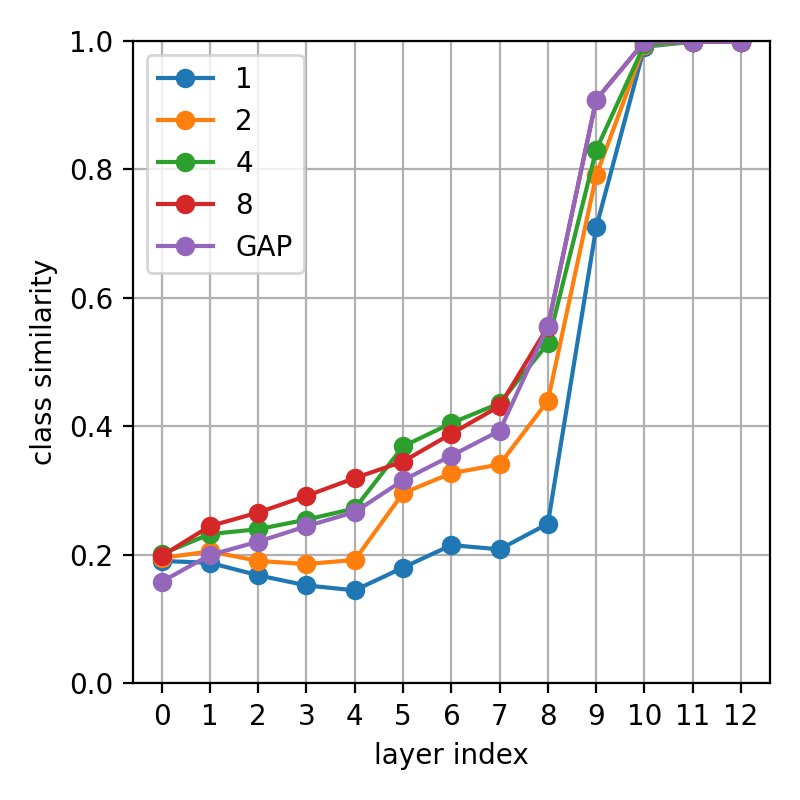}} 
    & 
    \subfloat[ImageNet64x64-50 \label{subfig:feature_dim_avgpool_CS_img64}]{\includegraphics[width=0.45\linewidth]{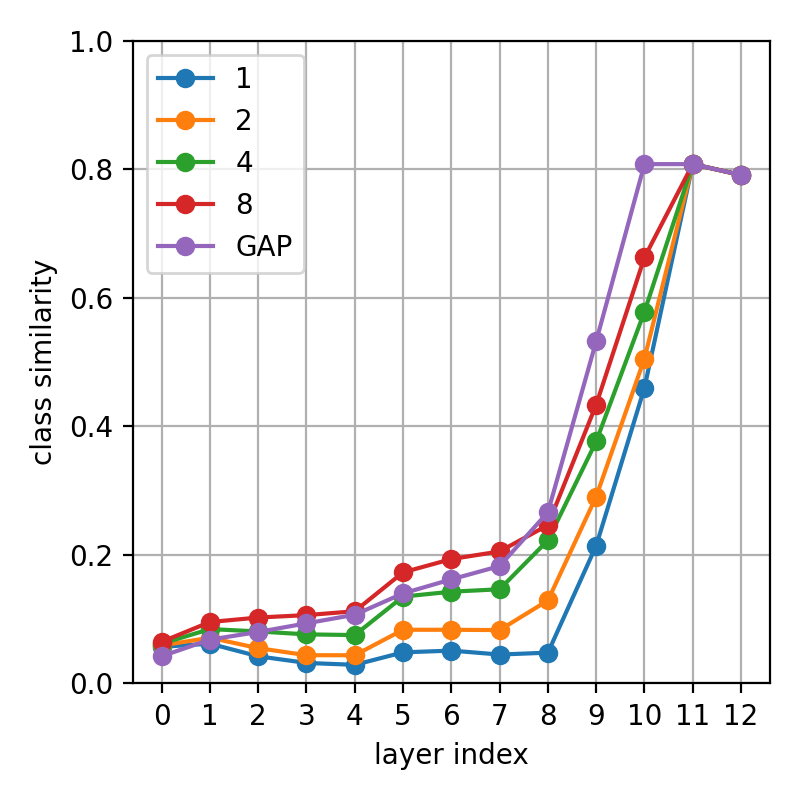}}
    \\
    \end{tabular}
    \caption{\small Multi-scale class similarity curve of a ResNet20 model trained on different datasets. }
    \label{fig:feature_dim_CS}
\end{figure}

In this section we investigate using a non-overlapping average pooling at 5 spatial scales $S$: identity, 2x, 4x, 8x, and global pooling. Each scale is used to generate a separate class similarity curve for two different datasets. The curves for each dataset is shown in Figure~\ref{fig:feature_dim_CS}. The x-axis for each figure shows the layer index within the ResNet20 model. Layer 0 is preprocessed dataset, layer 1 is the output of the ResNet neck, and layers 2-4, 5-7, and 8-10 are the outputs of each layer in the first, second, and third ResNet stages, respectively. Layer 10 is the output of global average pooling, and layer 11 is the classification logits.

\begin{figure*}
    \centering
    \includegraphics[width=0.95\textwidth]{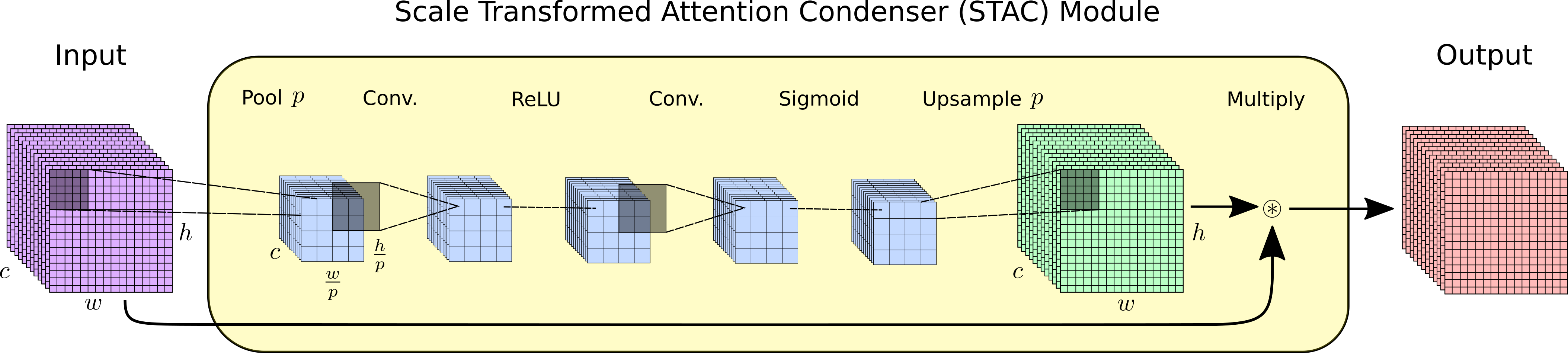}
    \caption{An illustration of a Scaled Transformed Attention Condenser (STAC) Module.}
    \label{fig:stac_module}
\end{figure*}

\textbf{Identity CS:} We now briefly review the CS curve when pooling is not applied (i.e., the blue curve). For the CIFAR10 identity CS curve there is an upward trending change as the layers get deeper. However, the upward trend is not monotonic; between layers 0 to 3 we see a small but non-trivial decrease in the CS curve before the the CS curve begins to trend upwards. We refer to this phenomenon as the \textbf{initial CS contraction}. There is a large increase in the observed CS between layer 8 to layer 10.  We refer to this change as the \textbf{CS cluster spike}. Interestingly, the CS curve up to the CS cluster spike is inversely correlated with the intrinsic dimensionality hunchbatch pattern observed in (Ansuini et al.)~\cite{ansuini2019intrinsic}, we leave this observation to future work to explore. Both of these trends are also present for Img64-50. Note that the CS for Img64-50 in the final layers of the model reaches a maximum of 0.81 despite achieving a training set accuracy of $100\%$, a clear indication that the model has over fit. 

\textbf{Average Pooling - CIFAR10:} When 2x average pooling is applied the CS curve see a small increase the first 5 layers. With 2x average pooling there is now only a negligible decrease in CS during the layers that originally experienced the initial CS contraction. During the second ResNet stage (i.e., layers 5, 6, and 7) the CS curve jumps. The subsequent transition between feature map sizes (i.e., between layer 7 and layer 8) also experiences a non-negligible increase in CS. The largest increase between layers still happens between layer 8 and layer 9. When the average pooling window is increased to 4x the CS for every layer continues to increase. In addition, the subtleties in the CS curve are diminished as the the curve between layer 0 and layer 8 begins to resemble a increasing line. However, the first 5 layers still have a smaller rate of increase, and a jump in CS still happens between layer 4 and layer 5. At 8x average pooling there is no longer a general increase in CS values. Despite this, the curve between layer 0 and layer 7 now more closely approximate line. The increased difference in CS between layer 4 and 5 which was introduced by 2x pooling and persisted by 4x pooling, is greatly diminished by 8x pooling. When global average pooling is applied the CS between layer 0 and layer 7 (i.e., layers that still have non-channel dimensions) all see a decrease in CS relative to the 8x pooling curve. In addition layer 0 now dips below the identity curve.  

\textbf{Average Pooling - Img64-50}: The results for applying average pooling to Img64-50 is shown in Figure~\ref{subfig:feature_dim_avgpool_CS_img64}. Like with CIFAR10, we see that the identity CS curve contains both the initial CS contraction, and the CS cluster spike. The main differences compared to CIFAR10 are the reduced magnitude of the CS curve, that the CS cluster spike happens over a larger range of layers, and that the spike doesn't approach 1.0. With 2x average pooling layers 0 to 4 experience minor change, layers 5 to 7 see a more substantial jump, and layer 8 sees the largest increase in CS out of all the layers. The increase in CS for subsequent layers diminishes until only negligible change is seen in layers 11 and 12. The 4x curve sees all CS values increase relative to the 2x CS curve, with the trend continuing between 4x and 8x. Taking a look at the global averaged pooled CS curve, one can see that the CS decreases relative to the 8x pooled curve for layers 0 to 7 but then under goes an increase between layers 8 to 10. In general, average pooling effects Img64-50 in a similar manner to CIFAR10, but where the changes induced by average pooling are delayed by one window size. That is, the shape of the CS curves tend to be more correlated with the feature map size of the pooled feature map then with the magnitude of the pooling window. In fact, when comparing pooled CIFAR10 curves and Img64-50 curves the average correlation coefficient between curves with equal average window size is 0.95, and with equal feature map size is 0.97. For context the average correlation coefficient between any two random CS curve pairs is 0.94. Another similarity between the global average pooling curves for both CIFAR10 and Img64-50 is that layer 0 to layer 7 closely approximates a line.

\section{Spatial Transformed Attention Condenser (STAC) Module}
\label{sec:spatial_attention_condensers}

In the last section ClassRepSim revealed to what degree different spatial resolutions of features result in more coherent class groupings. In general the larger the pooling window used on a sample's features the closer the sample is too samples from the same class for any given layer in a DNN. This observation holds for any pooling window size. However, peak class similarity for each layer often occurs with a pooling window that is smaller then the largest pooling window size. In this section we propose spatial transformed attention condenser (STAC) modules, a specific instantiation of attention condensers, which is designed to take advantage of the heightened class similarity observed at specific spatial resolutions. 

The proposed STAC module is constructed as follows. Given an input feature map $Y^i$, a condenser operation $Q^i=P(Y^i)$ spatially reduces the size of $Y^i$. Then a set of attention operations $A$ identify the important regions of $Q^i$ and produces an attention map $K^i=A(Q^i)$. An expansion layer $E$ spatially increases the attention map $K^i$ to a full-focus self-attention map $T^i=E(K^i)$ which matches $Y^i$'s spatial resolution. Finally, the full-focus self-attention map $T^i$ is element-wise multiplied with $Y^i$ to out the spatially transformed attention feature map $Y^j = Y^i \times T^i $. The STAC module is depicted in Figure~\ref{fig:stac_module}.

In this work $P$ is an average pooling operation where the stride size and pooling window are equal, $A$ is multi-layer sub-network with convolution, ReLU, convolution, sigmoid operations sequentially applied, and $E$ is a nearest neighbour upsampler. Conceptually the proposed STAC module follows a similar construction to a sequence and excitation network (SENet) but with the condenser operation having a limited window size and with the addition of an expansion operation to match the input feature map size. We hypothesize that the structural bias designed into the STAC module will allow a DNN to better make use of the intrinsic spatial features found at specific spatial resolutions within a dataset and will result in improved model performance with negligible increase in computation. 

We explore two different placement locations of the the STAC module, following the notation used in (Hu et al.)~\cite{hu2018squeeze}, 1) \textit{standard} placement where the STAC module is placed after the residual block but before the skip connection, and 2) \textit{post} placement where the STAC module is placed after the skip connection.

\section{Results}
\label{sec:results}

\begin{table*}
    \caption{Model Comparison - Top-1 Accuracy, FLOPs, Number of Parameters}
    \centering
    \setlength\tabcolsep{6.0pt} 
    \small
    \begin{tabular} { l | c  c  c | c  c  c | c  c  c }
        \toprule

        & \multicolumn{3}{c|}{\textbf{ResNet20 - CIFAR10}} & \multicolumn{3}{c|}{\textbf{ResNet34 - ImageNet64x64-50}} & \multicolumn{3}{c}{\textbf{ResNet34 - ImageNet64x64}} \\
        \multicolumn{1}{c|}{\textbf{Model}} & \textbf{Top-1 Acc.} & \textbf{FLOPs} & \textbf{Param.} & \textbf{Top-1 Acc.} & \textbf{FLOPs} & \textbf{Param.} & \textbf{Top-1 Acc.} & \textbf{FLOPs} & \textbf{Param.} \\
        \midrule
        Base~\cite{he2016deep} & 89.9 $\pm$ 0.2 & 41.5M & 272K & 76.8 $\pm$ 0.5 & 4.65B & 21.3M & 61.2 $\pm$ 0.1 & 4.65B & 21.8M\\
        Base - 2x-Deep & 90.7 $\pm$ 0.5 & 85.7M & 564K & 76.5 $\pm$ 0.5 & 9.50B & 44.0M & 62.4 $\pm$ 0.3 & 9.50B & 44.4M \\
        Base - $\sqrt{2}$x-Wide & \textbf{91.2} $\pm$ 0.2 & 78.5M & 522K & \textbf{77.4} $\pm$ 0.6 & 9.07B & 41.6M & \textbf{62.7} $\pm$ 0.1 & 9.19B & 42.6M \\
        \midrule
        SENet~\cite{hu2018squeeze} - Standard & 90.5 $\pm$ 0.2 & 41.7M & 305K & 77.6 $\pm$ 0.5 & 4.66B & 23.8M & 62.3 $\pm$ 0.2 & 4.66B & 24.3M\\
        BAM~\cite{park2018bam} - Standard & 90.5 $\pm$ 0.2 & 86.6M & 614K & 75.2 $\pm$ 0.7 & 9.76B & 47.7M & 62.0 $\pm$ 0.1 & 9.76B & 48.2M \\
        STAC - Standard (Ours) & \textbf{90.6} $\pm$ 0.2 & 42.3M & 563K & \textbf{77.7} $\pm$ 0.4 & 4.73B & 43.9M & \textbf{62.8} $\pm$ 0.1 & 4.73B & 44.4M\\
        \midrule
        SENet~\cite{hu2018squeeze} - Post & 90.4 $\pm$ 0.2 & 41.7M & 305K & 75.7 $\pm$ 0.5 & 4.66B & 23.8M & 61.7 $\pm$ 0.2 & 4.66B & 24.3M \\
        BAM~\cite{park2018bam} - Post & \textbf{90.7} $\pm$ 0.3 & 86.6M & 614K & * & 9.76B & 47.7M & * & 9.76B & 48.2M \\
        STAC - Post (Ours) & 90.4 $\pm$ 0.2 & 42.3M & 563K & \textbf{76.6} $\pm$ 0.4 & 4.73B & 43.9M & \textbf{62.2} $\pm$ 0.2 & 4.73B & 44.4M\\

        \bottomrule
    \end{tabular}    
    \label{tab:results}
    \\\footnotesize{*In these configurations the BAM module based models was unstable with the majority of the runs remaining near random performance.}\\
\end{table*}

\begin{figure}
    \centering
    \includegraphics[width=0.7\linewidth]{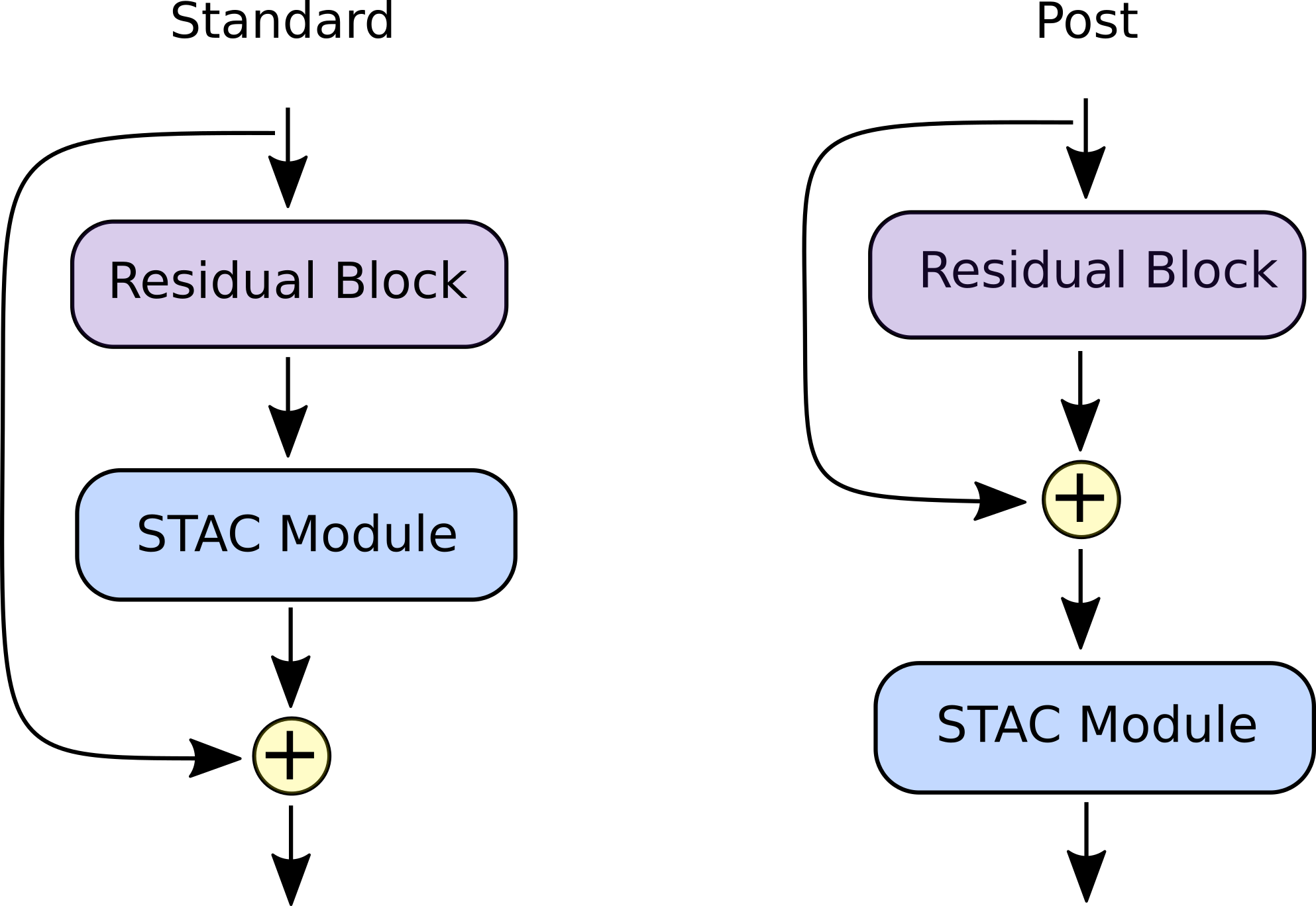}
    \caption{Examples of STAC module placement.}
    \label{fig:stac_module_placement}
\end{figure}

We now compare DNNs using the proposed STAC module to models using other self-attention modules as well as standard ResNet models adjusted to have a similar number of FLOPs or parameters. The other self-attention modules include SENet~\cite{hu2018squeeze} modules and BAM~\cite{park2018bam} modules (note that we use the module design but not the macro-architecture placement of these modules as presented in (Park et al.)~\cite{park2018bam}). Each of the three self-attention modules are tested in both the \textit{standard} and \textit{post} placement stategies described in (Hu et al.)~\cite{hu2018squeeze} (see Figure~\ref{fig:stac_module_placement}). For the STAC module we use a pooling window of 8 (since is provides the highest overall CS scores as observed in Figure~\ref{fig:feature_dim_CS}), and for the BAM module we use a dilation factor of 4 (for a similar receptive field as the STAC module) and a compression factor of 4. The standard ResNet models include a \textit{vanilla} ResNet model, a \textit{2x-deep} ResNet model, and a \textit{$\sqrt{2}$-wide} ResNet model. 

We compare the performance of these nine different models on three architecture-dataset pairs, including: ResNet20-Cifar10, ResNet34-Img64-50, and ResNet34-Img64, where the ResNet34 model is the normal size model from (He et al.)~\cite{he2016deep}, and Img64 is the complete ImageNet64x64 dataset~\cite{chrabaszcz2017downsampled}. The first model is trained for 100 epochs, a base learning rate of 0.1, and batch size of 128. The second model is trained for 100 epochs, a base learning rate of 0.01, and a batch size of 128. The third model is trained for 40 epochs, a base learning rate of 0.01, and a batch size of 256. The learning rates were selected to maximize the performance of the base ResNet model configuration for each dataset. All model configurations use the same set of crop, flip, rotation, colour jitter, grayscale, and blur augmentations. Results comparing the accuracy, FLOPs, and number of parameters of these models is shown in Figure~\ref{tab:results}. The accuracy results for each model is the average of 8 training runs.   

In all cases the use of the STAC-Standard module resulted in improved performance over the base model by $0.7\%$, $0.9\%$, and $1.6\%$ on the CIFAR10, Img64-50, and Img64 datasets, respectively, at a small increase to FLOPs but over double the number of parameters. For CIFAR10 both the 2x-deep and $\sqrt{2}$x-wide models outperformed the STAC-standard module, but both of these models require at least 1.85x more FLOPS with a similar number of parameters. However, for Img64-50 and Img64 the STAC-Standard module outperformed both variants of the base model with similar differences in FLOPs and parameter count. Amongst the self-attention modules using standard placement the STAC module outperformed the other two designs on all three datasets. On CIFAR10 and Img64-50 the STAC module narrowly outperformed SENet, while on Img64 the STAC module outperformed SENet by $0.5\%$ at the cost of $1.5\%$ addition FLOPs and almost 2x the number of parameters. 

In all but one case all module variants using post placement under performed standard placement, with the one exception being the BAM-post module trained on CIFAR10. In this instance the BAM-post module outperformed all other self-attention configurations but still under performed the $\sqrt{2}$x-wide base model variant. All other configurations using the BAM-post module were unstable during training using the selected training parameters. As a reminder, the training parameters used in this experiment were selected to optimize the respective base ResNet models. Overall, the highest performing model for CIFAR10 was ResNet20 $\sqrt{2}$x-wide, and for Img64-50 and Img64 was the ResNet34 models using the STAC module is standard placement.

\section{Ablation Results}
\label{sec:ablation}

Above it was shown that the STAC module can increase the performance of a variety of ResNet model sizes trained on datasets with different properties at the cost of a small increase in FLOPs and doubling of the number of parameters compared to standard ResNet models. In this section we perform a series of ablation tests on STAC based models. Each experiment is performed on a ResNet20 model trained on either CIFAR10 of Img64-50 using the same training parameters from Section~\ref{algo:ClassRepSim}. The tests in this section are explored using only the \textit{standard} location of the STAC module. Unless stated specified, the default configuration for all ResNet20 models in this section use STAC modules, standard placement, and with a condenser window size of 8. All results in this section are from 8 independent models.

\subsection{Self-Attention Receptive Field}
\label{sec:FOV}

The condenser operation (average pooling) used to reduce the spatial feature size for the down stream self-attention operations has the consequence of increasing the self-attention operation's receptive fields; namely the two convolutional operations in each STAC module. The 3x3 spatial shape of these kernels further expands the STAC module's receptive field such that the each feature in the output attention map is dependent on a larger portion of the original input feature map. Table~\ref{tab:kernel_size} shows the results of changing the spatial feature size of both the first convolution kernel $C1$ and the second convolution kernel $C2$ from a 3x3 kernel to a 1x1 kernel for all modules in the ResNet20 model.  

\begin{table}[ht]
    \caption{STAC Kernel Size - ResNet20}
    \centering
    \setlength\tabcolsep{6.0pt} 
    \small
    \begin{tabular} { l | c @{\hspace{1mm}} c | c c c }
        \toprule

        \textbf{Dataset} & \textbf{C1} & \textbf{C2} & \textbf{Top-1 Acc.} & \textbf{FLOPs} & \textbf{Params.} \\
        \midrule
        CIFAR10 & 1 & 1 & 90.3 $\pm$ 0.2 & 41.7M & 305K \\
        & 1 & 3 & 90.4 $\pm$ 0.1 & 42.0M & 434K \\
        & 3 & 1 & 90.6 $\pm$ 0.3 & 42.0M & 434K \\
        & 3 & 3 & 90.6 $\pm$ 0.2 & 42.3M & 563K \\
        \midrule
        Img64-50 & 1 & 1 & 73.3 $\pm$ 0.4 & 167M & 308K \\
        & 1 & 3 & 73.4 $\pm$ 0.5 & 168M & 437K \\
        & 3 & 1 & 73.6 $\pm$ 0.6 & 168M & 437K \\
        & 3 & 3 & 73.6 $\pm$ 0.5 & 169M & 566K \\
        \bottomrule
    \end{tabular}    
    \label{tab:kernel_size}
\end{table}

A kernel configuration of $(1,1)$ has marginally more FLOPs but the same number of parameters as a model using SENet modules and performs within $\pm 0.3\%$ of the respective SENet modules (see the respective GAP rows in Table~\ref{tab:condenser_size}). For both datasets the kernel configuration $(3,1)$ showed marginal performance benefits over $(1,3)$, with $(3,1)$ matching the average performance of configuration $(3,3)$ but with marginally higher standard variation. For both datasets the configurations $(1,3)$ and $(3,1)$ are attractive options to use in a model given the limited performance implications while significantly reducing the number of parameters. 

\subsection{Location of STAC Module}
\label{sec:stac_loc}

As shown in Section~\ref{sec:multi-scale-cs} the stages of a ResNet model have different effects on the increase (or decrease) of the observed class similarity. We now explore to what degree (if any) each stage of a ResNet model benefits from using the STAC module, as well as the associated computational costs of each configuation. Table~\ref{tab:STAC_location} shows the results for a normal ResNet20 model, when the layers in only a single stage of a  ResNet20 use the STAC module, and when all layers in a ResNet20 model use the STAC module. 

\begin{table}[ht]
    \caption{STAC Stage Location - ResNet20}
    \centering
    \setlength\tabcolsep{6.0pt} 
    \small
    \begin{tabular} { l | c @{\hspace{1mm}} c @{\hspace{1mm}} c | c c c }
        \toprule

        \textbf{Dataset} & \textbf{S1} & \textbf{S2} & \textbf{S3} & \textbf{Top-1 Acc.} & \textbf{FLOPs} & \textbf{Params.} \\
        \midrule
        CIFAR10 & & & & 89.9 $\pm$ 0.2 & 41.5M & 272K \\
        & $\times$ & & & 90.1 $\pm$ 0.2 & 41.8M & 286K \\
        &  & $\times$ &  & 90.4 $\pm$ 0.3 & 41.8M & 328K \\
        &  &  & $\times$ & 90.1 $\pm$ 0.2 & 41.8M & 494K \\
        & $\times$ & $\times$ & $\times$ & 90.6 $\pm$ 0.2 & 42.3M & 563K\\
        \midrule
        Img64-50 &  &  &  & 71.6 $\pm$ 0.5 & 166M & 275K \\
        & $\times$ &  &  & 71.7 $\pm$ 0.6 & 167M & 289K \\
        &  & $\times$ &  & 72.2 $\pm$ 0.7 & 167M & 330K \\
        &  &  & $\times$ & 73.0 $\pm$ 1.0 & 167M & 496K \\
        & $\times$ & $\times$ & $\times$ & 73.6 $\pm$ 0.5 & 169M & 566K\\

        \bottomrule
    \end{tabular}    
    \label{tab:STAC_location}
\end{table}

For CIFAR10 the STAC module provides a performance benefit when applied to all layers individually and has the most impact when applied to stage $S2$ falling $0.2\%$ short of the complete configuration. Stages $S1$ and $S3$ provided accuracy gains even through the $S3$ configuration requires significantly more parameters. Using STAC modules with Img64-50 shows a different pattern where the deeper the stage the more impact the STAC modules provides. Unlike with CIFAR10, when trained on Img64-50 no single stage configuration approaches the performance of using the STAC module throughout the model.

\subsection{Condenser Size}
\label{sec:condenser_size}

The size of the condenser window used in Section~\ref{sec:results} (i.e., $P=8$) was derived through observations made from the CS curves in Section~\ref{sec:class_similarity}. We now explore a wider range of condenser windows sizes to see how this choice effects the computational trade offs of using the STAC module. Table~\ref{tab:condenser_size} shows the result of changing the condenser window size from 1 (i.e., no pooling) to using global average pooling on the entire input feature map (i.e., resulting in SENet). 

\begin{table}[ht]
    \caption{STAC Condenser Size - ResNet20}
    \centering
    \setlength\tabcolsep{6.0pt} 
    \small
    \begin{tabular} { l | c | c c c }
        \toprule

        \textbf{Dataset} & \textbf{P} & \textbf{Top-1 Acc.} & \textbf{FLOPs} & \textbf{Params.} \\
        \midrule
        CIFAR10 & 1 & 90.7 $\pm$ 0.3 & 84.2M & 563K\\
        & 2 & 90.6 $\pm$ 0.2 & 52.3M & 563K \\
        & 4 & 90.5 $\pm$ 0.2 & 44.3M & 563K \\
        & 8 & 90.6 $\pm$ 0.2 & 42.3M & 563K \\
        & GAP & 90.5 $\pm$ 0.2 & 41.7M & 305K  \\
        \midrule
        Img64-50 & 1 & 73.1 $\pm$ 0.5 & 337M & 566K \\
        & 2 & 73.7 $\pm$ 0.2 & 209M & 566K \\
        & 4 & 73.6 $\pm$ 0.2 & 177M & 566K \\
        & 8 & 73.6 $\pm$ 0.5 & 169M & 566K \\
        & 16 & 73.6 $\pm$ 0.6 & 167M & 566K \\
        & GAP & 73.0 $\pm$ 0.5 & 167M & 308K  \\

        \bottomrule
    \end{tabular}    
    \label{tab:condenser_size}
\end{table}

Most STAC module windows sizes for both dataset perform near peak performance. For CIFAR10 all window sizes achieve within $0.2\%$ of peak performance, even though the number of FLOPs is significantly increased for smaller condenser window sizes (i.e., the feature maps in the STAC module are larger). The highest performing window size is $1$, which requires over double the number of FLOPs as using GAP pooling. However, for Img64-50 the top performing window sizes fall between 2 to 16, with the larger window sizes requiring less FLOPs but with greater variation in performance between runs. Both the smallest and largest window sizes under perform by at least $0.5\%$.

\subsection{Intervention Strategies}
\label{sec:STAC_strats}

It was shown in Section~\ref{sec:stac_loc} that the STAC module location is an important consideration in performance trade offs, and in Section~\ref{sec:condenser_size} that various condenser sizes provide comparable accuracy. We now consider three different strategies for selecting the condenser window size used for any given stage in a ResNet model. The First approach uses a \textit{greedy} strategy where we repeats the stage location experiment from Section~\ref{sec:stac_loc} but for every viable window size, then each stage is assigned the highest performing window size for each respective stage. The second strategy assigns the window size corresponding to the \textit{maximum CS} observed in Figure~\ref{fig:feature_dim_CS} for each respective dataset. The third strategy uses the largest possible uniform window size (i.e., the feature map spatial length in the final ResNet stage). The window size, accuracy, and FLOPs for these three strategies is shown in Table~\ref{tab:STAC_strategies}.

\begin{table}[ht]
    \caption{STAC Parameterization Selection Strategies - ResNet20}
    \centering
    \setlength\tabcolsep{6.0pt} 
    \small
    \begin{tabular} { l | c | c @{\hspace{1mm}} c @{\hspace{1mm}} c | c c c}
        \toprule

        \textbf{Dataset} & \textbf{Strategy} & \textbf{S1} & \textbf{S2} & \textbf{S3} & \textbf{Top-1 Acc.} & \textbf{FLOPs}   \\
        \midrule
        CIFAR10 & Greedy & 16 & 8 & 1 & 90.8 $\pm$ 0.2 & 56.0M  \\
        CIFAR10 & Max CS & 8 & 4 & 8 & 90.5 $\pm$ 0.2 & 43.0M \\
        CIFAR10 & Max Uni. & 8 & 8 & 8 & 90.6 $\pm$ 0.2 & 42.3M \\
        \midrule
        Img64-50 & Greedy & 32 & 4 & 2 & 73.5 $\pm$ 0.6 & 184M    \\
        Img64-50 & Max CS & 8 & 8 & 16 & 73.4 $\pm$ 0.4 & 168M    \\
        Img64-50 & Max Uni. & 16 & 16 & 16 & 73.5 $\pm$ 0.5 & 167M \\
        \bottomrule
    \end{tabular}    
    \label{tab:STAC_strategies}
\end{table}

Overall, each of the placement strategies results in similar performance. For CIFAR10 the greedy strategy produced the best accuracy, costing an addition 13M FLOPs, while the three strategies for Img64-50 demonstrated comparable performance. For both datasets the greedy strategy followed a similar progression of window sizes, with first stage using a the largest window size and getting progressively smaller as the stages progress. The maximum CS strategy favoured similar window sizes across all three stages for both datasets, with the last stage using the equivalent of global pooling.

\section{Discussion}

In this work we proposed ClassRepSim (Section~\ref{sec:class_similarity}) to study the class based characteristics of a dataset's latent space throughout a DNN at different spatial resolutions. From this analysis we identified that there exists a scale at which samples are most similar to other samples of the same class, that this peak scale changes throughout a model, and that the base spatial resolution of the feature maps is a key contributor to the optimal scale. Using the observations gained through ClassRepSim analysis we proposed STAC modules (Section~\ref{sec:spatial_attention_condensers}), a self-attention model designed to improve a DNN's ability to learn class based feature detectors. We demonstrated that this evidence based self-attention design results in improved performance in general over base models, out performs other self-attention modules in most cases, and who's performance gains required less additional FLOPs compared to larger variants of base models for similar performance gains (Section~\ref{sec:results}). 

We investigated a large assortment of STAC module parameterizations, including kernel sizes, placement of the STAC module, the size of the condenser window, and various strategies of hyperparameter selection (Section~\ref{sec:ablation}). Overall, we found that the performance gains from using the STAC module to be relatively robust in the selection of hyperparameters when the STAC module is used throughout a model. In general, to see desirable performance increases we recommend that STAC modules should be used throughout a model with condenser window sizes chosen with the aid of ClassRepSim. For CIFAR10 and Img64, this range is approximately between 4 to 16. Under these conditions our experiments demonstrate that STAC module is a viable building block for ResNet based classification DNNs. 

ClassRepSim shares similarities with other representational response metrics. As noted in Section~\ref{sec:class_similarity}, one such relation includes intrinsic dimensional of the feature space. Future work is required to identify how ClassRepSim can be used in conjungtion with other analytic methods to improve the repeatability of novel architecture improvements. For STAC modules additional experiments on a larger set of model-dataset combinations is required to prove out its generalizability and robustness to hyperparameter selection.

{\small
\bibliographystyle{ieee_fullname}
\bibliography{egbib}
}

\end{document}